\documentclass{article}

\usepackage{cite}

\usepackage[sort&compress,numbers]{natbib}

\usepackage[colorlinks=true,citecolor=blue,linkcolor=blue,urlcolor=blue]{hyperref}

\usepackage[preprint]{neurips_2024}
\usepackage{tabularx}
\usepackage{bm}
\usepackage{times}
\usepackage{latexsym}
\usepackage{colortbl}
\usepackage{float}
\usepackage{arydshln}
\usepackage{hyperref}
\usepackage[T1]{fontenc}

\usepackage[utf8]{inputenc}
\usepackage{enumitem}

\usepackage{microtype}

\usepackage{inconsolata}

\usepackage{graphicx}
\usepackage{url}
\usepackage{array}
\usepackage{amssymb} 
\usepackage{pifont} 
\newcommand{\xmark}{\ding{55}} 
\usepackage{multirow}
\usepackage{multicol}
\newcolumntype{P}[1]{>{\arraybackslash}p{#1}}
\newcolumntype{M}[1]{>{\centering\arraybackslash}m{#1}}
\usepackage{booktabs}
\usepackage{makecell}
\usepackage{amsmath}
\usepackage{xcolor}
\usepackage{wrapfig}
\title{Searching for Best Practices in Retrieval-Augmented Generation}

\author{Xiaohua Wang, \
{\bf Zhenghua Wang}, \
{\bf Xuan Gao}, \
{\bf Feiran Zhang}, \\
{\bf Yixin Wu}, \
{\bf Zhibo Xu}, \
{\bf Tianyuan Shi}, \
{\bf Zhengyuan Wang}, \
{\bf Shizheng Li}, \\
{\bf Qi Qian}, \
{\bf Ruicheng Yin}, \
{\bf Changze Lv}, \
{\bf Xiaoqing Zheng\thanks{Corresponding Author.}}, \
{\bf Xuanjing Huang} \\
School of Computer Science, Fudan University, Shanghai, China \\
Shanghai Key Laboratory of Intelligent Information Processing \\
\texttt{$\{$xiaohuawang22,zhenghuawang23$\}$@m.fudan.edu.cn}\\
\texttt{$\{$zhengxq,xjhuang$\}$@fudan.edu.cn}
}

\begin{document}

\maketitle

\begin{abstract}
Retrieval-augmented generation (RAG) techniques have proven to be effective in integrating up-to-date information, mitigating hallucinations, and enhancing response quality, particularly in specialized domains. 
While many RAG approaches have been proposed to enhance large language models through query-dependent retrievals, these approaches still suffer from their complex implementation and prolonged response times. 
Typically, a RAG workflow involves multiple processing steps, each of which can be executed in various ways. 
Here, we investigate existing RAG approaches and their potential combinations to identify optimal RAG practices. 
Through extensive experiments, we suggest several strategies for deploying RAG that balance both performance and efficiency.
Moreover, we demonstrate that multimodal retrieval techniques can significantly enhance question-answering capabilities about visual inputs and accelerate the generation of multimodal content using a ``retrieval as generation'' strategy.
Resources are available at \url{https://github.com/FudanDNN-NLP/RAG}.

\end{abstract}

\begin{figure*}[!t]
  \centering
  \includegraphics[width=0.95\linewidth]{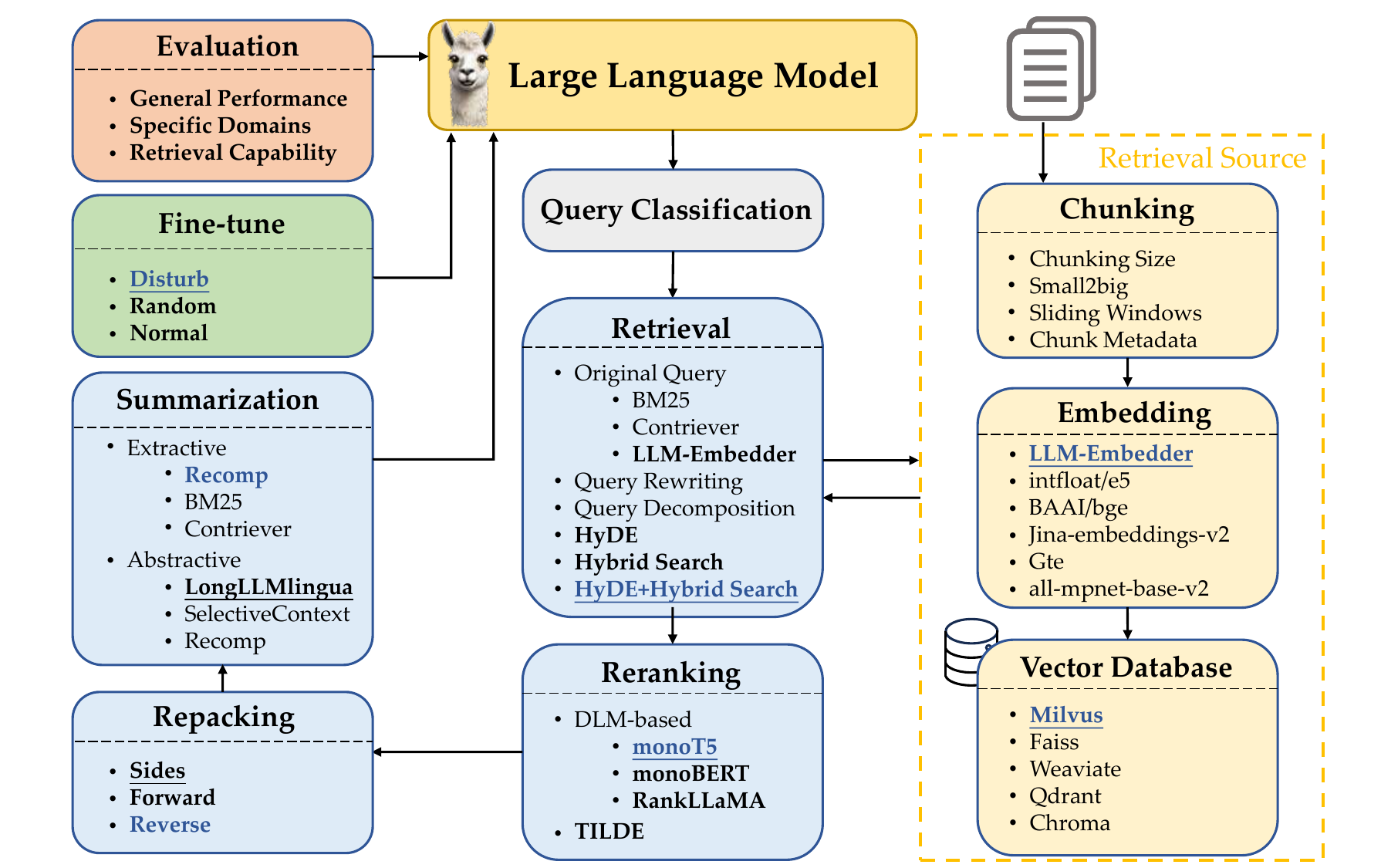}
  \caption{\label{fig:workflow} Retrieval-augmented generation workflow. 
  This study investigates the contribution of each component and provides insights into optimal RAG practices through extensive experimentation. The optional methods considered for each component are indicated in \textbf{bold} fonts, while the methods \underline{underlined} indicate the default choice for individual modules. The methods indicated in {\color{blue} blue} font denote the best-performing selections identified empirically.}
   
\end{figure*}

\section{Introduction}

Generative large language models are prone to producing outdated information or fabricating facts, although they were aligned with human preferences by reinforcement learning~\cite{ouyang2022training} or lightweight alternatives~\cite{rafailov2023direct,zhao2023slic,yuan2023rrhf,liu2023aligning}.  
Retrieval-augmented generation (RAG) techniques address these issues by combining the strengths of pretraining and retrieval-based models, thereby providing a robust framework for enhancing model performance~\cite{gao2023retrieval}. 
Furthermore, RAG enables rapid deployment of applications for specific organizations and domains without necessitating updates to the model parameters, as long as query-related documents are provided.


Many RAG approaches have been proposed to enhance large language models (LLMs)  through query-dependent retrievals~\cite{li2022survey, cai2022recent, gao2023retrieval}.
A typical RAG workflow usually contains multiple intervening processing steps: query classification (determining whether retrieval is necessary for a given input query), retrieval (efficiently obtaining relevant documents for the query), reranking (refining the order of retrieved documents based on their relevance to the query), repacking (organizing the retrieved documents into a structured one for better generation), summarization (extracting key information for response generation from the repacked document and eliminating redundancies) modules.
Implementing RAG also requires decisions on the ways to properly split documents into chunks, the types of embeddings to use for semantically representing these chunks, the choice of vector databases to efficiently store feature representations, and the methods for effectively fine-tuning LLMs (see Figure \ref{fig:workflow}). 

What adds complexity and challenge is the variability in implementing each processing step. 
For example, in retrieving relevant documents for an input query, various methods can be employed. 
One approach involves rewriting the query first and using the rewritten queries for retrieval~\cite{ma2023query}. 
Alternatively, pseudo-responses to the query can be generated first, and the similarity between these pseudo-responses and the backend documents can be compared for retrieval~\cite{gao2022precise}. 
Another option is to directly employ embedding models, typically trained in a contrastive manner using positive and negative query-response pairs~\cite{wang2022text, bge_embedding}.
The techniques chosen for each step and their combinations significantly impact both the effectiveness and efficiency of RAG systems. 
To the best of our knowledge, there has been no systematic effort to pursue the optimal implementation of RAG, particularly for the entire RAG workflow.



In this study, we aim to identify the best practices for RAG through extensive experimentation.
Given the infeasibility of testing all possible combinations of these methods, we adopt a three-step approach to identify optimal RAG practices.
First, we compare representative methods for each RAG step (or module) and select up to three of the best-performing methods. 
Next, we evaluate the impact of each method on the overall RAG performance by testing one method at a time for an individual step, while keeping the other RAG modules unchanged.
This allows us to determine the most effective method for each step based on its contribution and interaction with other modules during response generation.
Once the best method is chosen for a module, it is used in subsequent experiments. 
Finally, we empirically explore a few promising combinations suitable for different application scenarios where efficiency might be prioritized over performance, or vice versa. Based on these findings, we suggest several strategies for deploying RAG that balance both performance and efficiency.


The contributions of this study are three-fold:

\begin{itemize}[left=0pt]
\setlength{\parsep}{0pt}
\setlength{\parskip}{0pt}
\item Through extensive experimentation, we thoroughly investigated existing RAG approaches and their combinations to identify and recommend optimal RAG practices. 
\item We introduce a comprehensive framework of evaluation metrics and corresponding datasets to comprehensively assess the performance of retrieval-augmented generation models, covering general, specialized (or domain-specific), and RAG-related capabilities.
\item We demonstrate that the integration of multimodal retrieval techniques can substantially improve question-answering capabilities on visual inputs and speed up the generation of multimodal content through a strategy of ``retrieval as generation''. 
\end{itemize}

\section{Related Work}

Ensuring the accuracy of responses generated by Large Language Models (LLMs) such as ChatGPT~\cite{gpt4} and LLaMA~\cite{touvron2023llama} is essential. 
However, simply enlarging model size does not fundamentally address the issue of hallucinations~\cite{zhang2023siren, wang2023hallucination}, especially in knowledge-intensive tasks and specialized domains. 
Retrieval-augmented generation (RAG) addresses these challenges by retrieving relevant documents from external knowledge bases, providing accurate, real-time, domain-specific context to LLMs~\cite{gao2023retrieval}.
Previous works have optimized the RAG pipeline through query and retrieval transformations, enhancing retriever performance, and fine-tuning both the retriever and generator. 
These optimizations improve the interaction between input queries, retrieval mechanisms, and generation processes, ensuring the accuracy and relevance of responses.

\subsection{Query and Retrieval Transformation}

Effective retrieval requires queries accurate, clear, and detailed. 
Even when converted into embeddings, semantic differences between queries and relevant documents can persist. 
Previous works have explored methods to enhance query information through query transformation, thereby improving retrieval performance.  
For instance, Query2Doc~\cite{wang2023query2doc} and HyDE~\cite{gao2022precise} generate pseudo-documents from original queries to enhance retrieval, while TOC~\cite{kim2023tree} decomposes queries into subqueries, aggregating the retrieved content for final results.

Other studies have focused on transforming retrieval source documents. 
LlamaIndex~\cite{Liu_LlamaIndex_2022} provides an interface to generate pseudo-queries for retrieval documents, improving matching with real queries. 
Some works employ contrastive learning to bring query and document embeddings closer in semantic space~\cite{zhang2023retrieve,bge_embedding,li2023towards}.
Post-processing retrieved documents is another method to enhance generator output, with techniques like hierarchical prompt summarization~\cite{jiang2023llmlingua} and using abstractive and extractive compressors~\cite{xu2023recomp} to reduce context length and remove redundancy~\cite{wang2023learning}.

\subsection{Retriever Enhancement Strategy}

Document chunking and embedding methods significantly impact retrieval performance. 
Common chunking strategies divide documents into chunks, but determining optimal chunk length can be challenging. 
Small chunks may fragment sentences, while large chunks might include irrelevant context. 
LlamaIndex~\cite{Liu_LlamaIndex_2022} optimizes the chunking method like Small2Big and sliding window.
Retrieved chunks can be irrelevant and numbers can be large, so reranking is necessary to filter irrelevant documents.
A common reranking approach employs deep language models such as BERT~\cite{nogueira2019multi}, T5~\cite{nogueira2020document}, or LLaMA~\cite{ma2023fine}, which requires slow inference steps during reranking but grants better performance. 
TILDE~\cite{zhuang2021tilde, zhuang2021fast} achieves efficiency by precomputing and storing the likelihood of query terms, ranking documents based on their sum.

\subsection{Retriever and Generator Fine-tuning}
 
Fine-tuning within the RAG framework is crucial for optimizing both retrievers and generators.
Some research focuses on fine-tuning the generator to better utilize retriever context~\cite{Luo2023SAILSI, Zhang2024RAFTAL, Liu2024ChatQASG}, ensuring faithful and robust generated content.
Others fine-tune the retriever to learn to retrieve beneficial passages for the generator~\cite{Izacard2022FewshotLW, Zhang2024ARL2AR,shi2023replug}. 
Holistic approaches treat RAG as an integrated system, fine-tuning both retriever and generator together to enhance overall performance~\cite{Guu2020REALMRL, Lin2023RADITRD, Zamani2024StochasticRE}, despite increased complexity and integration challenges.

Several surveys have extensively discussed current RAG systems, covering aspects like text generation~\cite{li2022survey, cai2022recent}, integration with LLMs~\cite{gao2023retrieval, huang2024survey}, multimodal~\cite{zhao2023retrieving}, and AI-generated content~\cite{zhao2024retrieval}. 
While these surveys provide comprehensive overviews of existing RAG methodologies, selecting the appropriate algorithm for practical implementation remains challenging. 
In this paper, we focus on best practices for applying RAG methods, advancing the understanding and application of RAG in LLMs.

\begin{figure*}[t]
    \centering
    \includegraphics[width=0.95\linewidth]{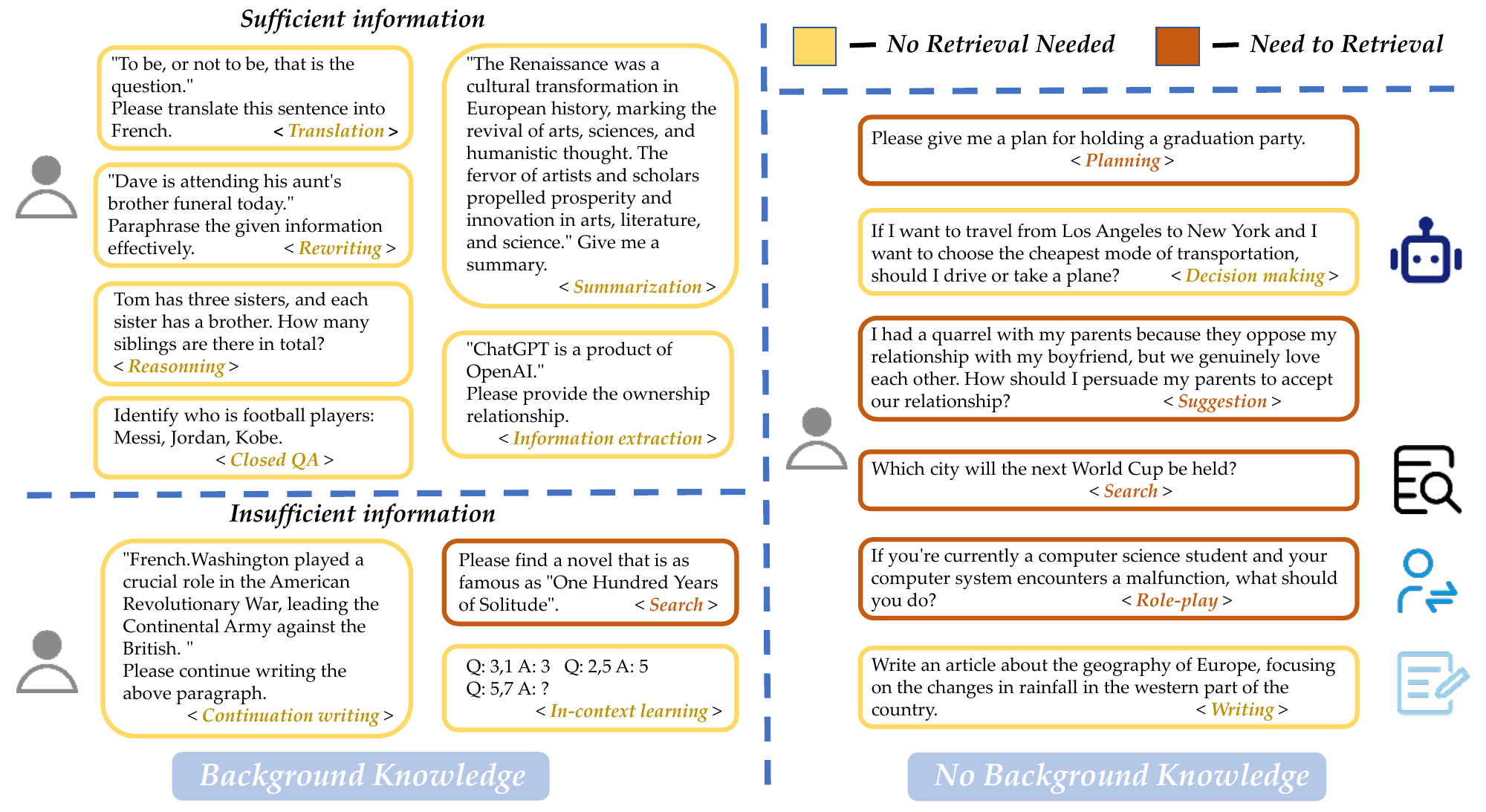}
    \caption{\label{fig:task_cls} Classification of retrieval requirements for different tasks. 
    In cases where information is not provided, we differentiate tasks based on the functions of the model.}
\end{figure*}

\section{RAG Workflow}
\label{workflow}
 
In this section, we detail the components of the RAG workflow. For each module, we review commonly used approaches and select the default and alternative methods for our final pipeline. Section \ref{section4} will discuss best practices. Figure \ref{fig:workflow} presents the workflow and methods for each module. Detailed experimental setups, including datasets, hyperparameters, and results are provided in Appendix \ref{sec:appendix}.

\subsection{Query Classification}
 
Not all queries require retrieval-augmented due to the inherent capabilities of LLMs. 
While RAG can enhance information accuracy and reduce hallucinations, frequent retrieval can increase response time. 
Therefore, we begin by classifying queries to determine the necessity of retrieval. 
Queries requiring retrieval proceed through the RAG modules; others are handled directly by LLMs.

Retrieval is generally recommended when knowledge beyond the model's parameters is needed. 
However, the necessity of retrieval varies by task. 
For instance, an LLM trained up to 2023 can handle a translation request for ``\textit{Sora was developed by OpenAI}'' without retrieval. 
Conversely, an introduction request for the same topic would require retrieval to provide relevant information.

Therefore, we propose classifying tasks by type to determine if a query needs retrieval.
We categorize
\begin{wraptable}{r}{0.5\textwidth}
    \centering
    \setlength{\tabcolsep}{3pt}
    \small
    \begin{tabular}{lcccc}
        \toprule
        \multirow{2}{*}{\textbf{Model}} & & \multicolumn{2}{c}{\textbf{Metrics}} \\  \cmidrule(lr){2-5}  
        & Acc & Prec & Rec & F1 \\
        \midrule
         \makecell{BERT-base-multilingual} & 0.95 & 0.96 & 0.94 & 0.95 \\
        \bottomrule
    \end{tabular}
    \caption{\label{tab:task cls}Results of the Query Classifier.}
\end{wraptable}
15 tasks based on whether they provide sufficient information, with specific tasks and examples
illustrated in Figure \ref{fig:task_cls}. 
For tasks entirely based on user-given information, we denote as \textbf{``sufficient''}, which need not retrieval; otherwise, we denote as \textbf{``insufficient''}, and retrieval may be necessary.
We train a classifier to automate this decision-making process.
Experimental details are presented in Appendix \ref{Query Classification Appendix}.
Section \ref{section4} explores the impact of query classification on the workflow, comparing scenarios with and without classification.

\subsection{Chunking}
\label{chunk section}
Chunking documents into smaller segments is crucial for enhancing retrieval precision and avoiding length issues in LLMs. 
This process can be applied at various levels of granularity, such as token, sentence, and semantic levels.
\begin{itemize}[left=0pt]
\setlength{\parsep}{0pt}
\setlength{\parskip}{0pt}
\item \textbf{Token-level Chunking} is straightforward but may split sentences, affecting retrieval quality.
\item \textbf{Semantic-level Chunking} uses LLMs to determine breakpoints, context-preserving but time-consuming.
\item \textbf{Sentence-level Chunking} balances preserving text semantics with simplicity and efficiency.
 
\end{itemize}
 
In this study, we use \underline{\textbf{sentence-level chunking}}, balancing simplicity and semantic preservation. 
We examine chunking from four dimensions.

\begin{table*}[t]
  \centering
  \setlength{\tabcolsep}{3pt}
  \small
  \begin{tabular}{lcccccccc}
    \hline
    \multirow{2}{*}{\raisebox{-2mm}{\textbf{Embedding Model}}} & \multicolumn{6}{c}{\raisebox{-1mm}{\textbf{namespace-Pt/msmarco}}} \\
    \cmidrule(lr){2-7}
    & MRR@1 & MRR@10 & MRR@100 & R@1  & R@10 & R@100 \\
    \hline
    BAAI/LLM-Embedder~\cite{zhang2023retrieve} & \underline{$24.79$} & \underline{$37.58$} & \underline{$38.62$} & \underline{$24.07$} & \bm{$66.45$} & \bm{$90.75$} \\
    BAAI/bge-base-en-v1.5~\cite{bge_embedding} & $23.34$ & $35.80$ & $36.94$ & $22.63$ & $64.12$ & $90.13$ \\
    BAAI/bge-small-en-v1.5~\cite{bge_embedding} & $23.27$ & $35.78$ & $36.89$ & $22.65$ & $63.92$ & $89.80$ \\
    BAAI/bge-large-en-v1.5~\cite{bge_embedding} & $24.63$ & $37.48$ & $38.59$ & $23.91$ & $65.57$ & $90.60$ \\
    BAAI/bge-large-en~\cite{bge_embedding} & \bm{$24.84$} & \bm{$37.66$} & \bm{$38.73$} & \bm{$24.13$} & $66.09$ & $90.64$ \\
    BAAI/bge-small-en~\cite{bge_embedding} & $23.28$ & $35.79$ & $36.91$ & $22.62$ & $63.96$ & $89.67$ \\
    BAAI/bge-base-en~\cite{bge_embedding} & $23.47$ & $35.94$ & $37.07$ & $22.73$ & $64.17$ & $90.14$ \\
    Alibaba-NLP/gte-large-en-v1.5~\cite{li2023towards} & $8.93$ & $15.60$ & $16.71$ & $8.67$ & $32.28$ & $60.36$ \\
    thenlper/gte-base~\cite{li2023towards} & $7.42$ & $13.23$ & $14.30$ & $7.21$ & $28.27$ & $56.20$ \\
    thenlper/gte-small~\cite{li2023towards} & $7.97$ & $14.81$ & $15.95$ & $7.71$ & $32.07$ & $61.08$ \\
    jinaai/jina-embeddings-v2-small-en~\cite{gunther2023jina} & $8.07$ & $15.02$ & $16.12$ & $7.87$ & $32.55$ & $60.36$ \\
    intfloat/e5-small-v2~\cite{wang2022text} & $10.04$ & $18.23$ &$19.41$ & $9.74$ & $38.92$ & $68.42$ \\
    intfloat/e5-large-v2~\cite{wang2022text} & $9.58$ & $17.94$ & $19.03$ & $9.35$ & $39.00$ & $66.11$ \\
    sentence-transformers/all-mpnet-base-v2 & $5.80$ & $11.26$ & $12.26$ & $5.66$ & $25.57$ & $50.94$ \\
    \hline
  \end{tabular}
  \caption{\label{tab:embedding model}
    Results for different embedding models on namespace-Pt/msmarco.
  }
  \vspace{-3mm}
\end{table*}

\subsubsection{Chunk Size}
Chunk size significantly impacts performance. 
Larger chunks provide more context, enhancing comprehension but increasing process time.
Smaller chunks improve retrieval recall and reduce time but may lack sufficient context.

Finding the optimal chunk size involves a balance between some metrics such as faithfulness, and relevancy. 
Faithfulness measures whether the response is hallucinated or matches the retrieved texts. 
\begin{wraptable}{r}{0.5\textwidth}
  \centering
  \small
  \setlength{\tabcolsep}{5pt}
  \begin{tabular}{lcc}
    \hline
    \multirow{2}{*}{\raisebox{-5mm}{\textbf{\makecell{Chunk Size}}}} & \multicolumn{2}{c}{\raisebox{-1mm}{\textbf{lyft\textunderscore2021}}} \\
    \cmidrule(lr){2-3}
    & \multicolumn{1}{c}{\makecell{Average \\Faithfulness}}&  \multicolumn{1}{c}{\makecell{Average \\Relevancy}} \\
    \hline
    $2048$     &      $80.37$        &       $91.11$ \\
    $1024$         &      $94.26$        &      $95.56$ \\
    $512$ &   \bm{$97.59$}  &      $97.41$ \\
    $256$ & $97.22$    &      \bm{$97.78$} \\
    $128$         &      $95.74$        &       $97.22$ \\
    \hline
  \end{tabular}
  \caption{Comparison of different chunk sizes. }
  \label{tab:chunk size}
\end{wraptable}
Relevancy measures whether the retrieved texts and responses match queries. 
We use the evaluation module of LlamaIndex~\cite{llamaindex} to calculate the metrics above.  
For embedding, we use the \texttt{text-embedding-ada-002}\footnote{\url{https://platform.openai.com/docs/guides/embeddings/embedding-models}} model, which supports long input length. We choose  \texttt{zephyr-7b-alpha}\footnote{\url{https://huggingface.co/HuggingFaceH4/zephyr-7b-alpha}} and  \texttt{gpt-3.5-turbo}\footnote{\url{https://www.openai.com/}} as generation model and evaluation model respectively. The size of the chunk overlap is 20 tokens. First sixty pages of the document \texttt{lyft\textunderscore2021\footnote{\url{https://raw.githubusercontent.com/run-llama/llama_index/main/docs/docs/examples/data/10k/lyft_2021.pdf}}} are used as corpus, then prompting LLMs to generate about one hundred and seventy queries according to chosen corpus. The impact of different chunk sizes is shown in Table \ref{tab:chunk size}.

\subsubsection{Chunking Techniques}
Advanced techniques such as small-to-big and sliding window improve retrieval quality by organizing chunk block relationships. 
Small-sized blocks are used to match queries, and larger blocks that include the small ones along with contextual information are returned.

To demonstrate the effectiveness of advanced chunking techniques, we use the LLM-Embedder~\cite{zhang2023retrieve} model as an embedding model. The smaller chunk size is $175$ tokens, the larger chunk size is $512$ tokens and the chunk overlap is $20$ tokens. Techniques like small-to-big and sliding window improve retrieval quality by maintaining context and ensuring relevant information is retrieved. Detailed results are shown in Table \ref{tab:chunk skill}.

\subsubsection{Embedding Model Selection}
Choosing the right embedding model is crucial for effective semantic matching of queries and chunk blocks.
We use the evaluation module of \texttt{FlagEmbedding}\footnote{\url{https://github.com/FlagOpen/FlagEmbedding}}
which uses the dataset 
\begin{wraptable}{r}{0.5\textwidth}
  \centering
  \small
  \setlength{\tabcolsep}{5pt}
  \begin{tabular}{lcc}
    \hline
    \multirow{2}{*}{\raisebox{-5mm}{\textbf{Chunk Skill}}} & \multicolumn{2}{c}{\raisebox{-1mm}{\textbf{lyft\textunderscore2021}}} \\
    \cmidrule(lr){2-3}
    & \multicolumn{1}{c}{\makecell{Average \\Faithfulness}}&  \multicolumn{1}{c}{\makecell{Average \\Relevancy}} \\
    \hline
    Original     &      $95.74$        &      $95.37$ \\
    small2big         &      $96.67$        &       $95.37$ \\
    sliding window & \bm{$97.41$}    &      \bm{$96.85$} \\
    \hline
  \end{tabular}
  \caption{Comparison of different chunk skills.}
  \label{tab:chunk skill}
\end{wraptable}
\texttt{namespace-Pt/msmarco}\footnote{\url{https://huggingface.co/datasets/namespace-Pt/msmarco}} as queries and dataset \texttt{namespace-Pt/msmarco-corpus}\footnote{\url{https://huggingface.co/datasets/namespace-Pt/msmarco-corpus}} as corpus to choose the appropriate open source embedding model. 
As shown in Table \ref{tab:embedding model}, LLM-Embedder~\cite{zhang2023retrieve} achieves comparable results with BAAI/bge-large-en~\cite{bge_embedding}, however, the size of the former is three times smaller than that of the latter.
Thus, we select the \underline{\textbf{LLM-Embedder}}~\cite{zhang2023retrieve} for its balance of performance and size.

\subsubsection{Metadata Addition}
Enhancing chunk blocks with metadata like titles, keywords, and hypothetical questions can improve retrieval, provide more ways to post-process retrieved texts, and help LLMs better understand retrieved information.
A detailed study on metadata inclusion will be addressed in future work.

\subsection{Vector Databases}
Vector databases store embedding vectors with their metadata, enabling efficient retrieval of documents relevant to queries through various indexing and approximate nearest neighbor (ANN) methods.

To select an appropriate vector database for our research, we evaluated several options based on four key criteria: multiple index types, billion-scale vector support, hybrid search, and cloud-native 
\begin{wraptable}{r}{0.5\textwidth}
\small
\centering
\setlength{\tabcolsep}{3pt}
    \begin{tabular}{l|c|c|c|c}
        \hline
        \textbf{Database} & \makecell{\textbf{Multiple} \\ \textbf{Index Type}} & \makecell{\textbf{Billion-} \\ \textbf{Scale}} & \makecell{\textbf{Hybrid} \\ \textbf{Search}} & \makecell{\textbf{Cloud-} \\ \textbf{Native}} \\
        \hline
        Weaviate & \makecell{\xmark} & \makecell{\xmark} & \makecell{\checkmark} & \makecell{\checkmark} \\
        Faiss & \makecell{\checkmark} & \makecell{\xmark} & \makecell{\xmark} & \makecell{\xmark} \\
        Chroma & \makecell{\xmark} & \makecell{\xmark} & \makecell{\checkmark} & \makecell{\checkmark} \\
        Qdrant & \makecell{\xmark} & \makecell{\checkmark} & \makecell{\checkmark} & \makecell{\checkmark} \\
        \textbf{Milvus} & \makecell{\checkmark} & \makecell{\checkmark} & \makecell{\checkmark} & \makecell{\checkmark} \\
        \hline
    \end{tabular}
    \caption{Comparison of Various Vector Databases}
    \label{tab:vector-databases}
    \vspace{-3mm}
\end{wraptable}
capabilities. 
These criteria were chosen for their impact on flexibility, scalability, and ease of deployment in modern, cloud-based infrastructures.
Multiple index types provide the flexibility to optimize searches based on different data characteristics and use cases. Billion-scale vector support is crucial for handling large datasets in LLM applications.
Hybrid search combines vector search with traditional keyword search, enhancing retrieval accuracy.
Finally, cloud-native capabilities ensure seamless integration, scalability, and management in cloud environments.
Table \ref{tab:vector-databases} presents a detailed comparison of five open-source vector databases: \textbf{Weaviate}, \textbf{Faiss}, \textbf{Chroma}, \textbf{Qdrant}, and \textbf{Milvus}.

Our evaluation indicates that \underline{\textbf{Milvus}} stands out as the most comprehensive solution among the databases evaluated, meeting all the essential criteria and outperforming other open-source options.

\begin{table*}[t]
  \centering
  \setlength{\tabcolsep}{3pt}
  \scriptsize
  \begin{tabular}{lcccccccccc}
    \hline
    \multirow{3}{*}{\textbf{Method}} & \multicolumn{5}{c}{\raisebox{-1mm}{\textbf{TREC DL19}}} & \multicolumn{5}{c}{\raisebox{-1mm}{\textbf{TREC DL20}}} \\
    \cmidrule(lr){2-6} \cmidrule(lr){7-11}
    & mAP & nDCG@10 & R@50 & R@1k & Latency & mAP & nDCG@10 & R@50 & R@1k & Latency\\
    \hline
    \textit{unsupervised} \\
    BM25 & $30.13$ & $50.58$ & $38.32$ & $75.01$ & \bm{$0.07$} & $28.56$ & $47.96$ & $46.18$ & $78.63$ & \bm{$0.29$} \\
    Contriever & $23.99$ & $44.54$ & $37.54$ & $74.59$ & $3.06$ & $23.98$ & $42.13$ & $43.81$ & $75.39$ & $0.98$ \\
    \hline
    \textit{supervised} \\
    LLM-Embedder & $44.66$ & $70.20$ & $49.06$ & $84.48$ & \underline{$2.61$} & $45.60$ & $68.76$ & $61.36$ & $84.41$ & \underline{$0.71$} \\
    \hspace{1em}+ Query Rewriting & $44.56$ & $67.89$ & $51.45$ & $85.35$ & $7.80$ & $45.16$ & $65.62$ & $59.63$ & $83.45$ & $2.06$ \\
    \hspace{1em}+ Query Decomposition & $41.93$ & $66.10$ & $48.66$ & $82.62$ & $14.98$ & $43.30$ & $64.95$ & $57.74$ & $84.18$ & $2.01$ \\
    \hspace{1em}+ HyDE & \underline{$50.87$} & \bm{$75.44$} & \underline{$54.93$} & {$88.76$} & $7.21$ & \underline{$50.94$} & \bm{$73.94$} & {$63.80$} & {$88.03$} & $2.14$ \\
    \hspace{1em}+ Hybrid Search & $47.14$ & $72.50$ & $51.13$ & \underline{$89.08$} & $3.20$ & $47.72$ & $69.80$ & \underline{$64.32$} & \underline{$88.04$} & $0.77$ \\
    \hspace{1em}+ HyDE + Hybrid Search & \bm{$52.13$} & \underline{$73.34$} & \bm{$55.38$} & \bm{$90.42$} & $11.16$ & \bm{$53.13$} & \underline{$72.72$} & \bm{$66.14$} & \bm{$90.67$} & $2.95$ \\
    \hline
  \end{tabular}
  \caption{\label{tab:quey transformation methods}
    Results for different retrieval methods on TREC DL19/20. The best result for each method is made bold and the second is underlined.
  }
\end{table*}

\subsection{Retrieval Methods}
 
Given a user query, the retrieval module selects the top-$k$ relevant documents from a pre-built corpus based on the similarity between the query and the documents.
The generation model then uses these documents to formulate an appropriate response to the query. 
However, original queries often underperform due to poor expression and lack of semantic information~\cite{gao2023retrieval}, negatively impacting the retrieval process.
To address these issues, we evaluated three query transformation methods using the LLM-Embedder recommended in Section \ref{chunk section} as the query and document encoder:

\begin{itemize}[left=0pt]
\setlength{\parsep}{0pt}
\setlength{\parskip}{0pt}
 
\item \textbf{Query Rewriting}: Query rewriting refines queries to better match relevant documents.
Inspired by the Rewrite-Retrieve-Read framework~\cite{ma2023query}, we prompt an LLM to rewrite queries to enhance performance.
\item \textbf{Query Decomposition}: This approach involves retrieving documents based on sub-questions derived from the original query, which is more complex to comprehend and handle.
\item \textbf{Pseudo-documents Generation}: This approach generates a hypothetical document based on the user query and uses the embedding of hypothetical answers to retrieve similar documents. One notable implement is HyDE~\cite{gao2022precise},
 
\end{itemize}

Recent studies, such as~\citep{sawarkar2024blended}, indicate that combining lexical-based search with vector search significantly enhances performance.
In this study, we use BM25 for sparse retrieval and Contriever~\cite{izacard2021unsupervised}, an unsupervised contrastive encoder, for dense retrieval, serving as two robust baselines based on~\citet{thakur2021beir}. 

\subsubsection{Results for different retrieval methods}
We evaluated the performance of different search methods on the TREC DL 2019 and 2020 passage ranking datasets.
The results presented in Table \ref{tab:quey transformation methods} show that supervised methods significantly outperformed unsupervised methods. Combining with HyDE and hybrid search, LLM-Embedder achieves the highest scores.
However, query rewriting and query decomposition did not enhance retrieval performance as effectively.
Considering the best performance and tolerated latency, we recommend \underline{\textbf{Hybrid Search with HyDE}} as the default retrieval method. Taking efficiency into consideration, \textbf{Hybrid Search} combines sparse retrieval (BM25) and dense retrieval (Original embedding) and achieves notable performance with relatively low latency.

\begin{table*}[t]
  \centering
  \setlength{\tabcolsep}{3pt}
  \scriptsize
  \begin{tabular}{lcccccccccc}
    \hline
     \multirow{3}{*}{\textbf{Configuration}} & \multicolumn{5}{c}{\raisebox{-1mm}{\textbf{TREC DL19}}} & \multicolumn{5}{c}{\raisebox{-1mm}{\textbf{TREC DL20}}} \\
    \cmidrule(lr){2-6} \cmidrule(lr){7-11}
    & mAP & nDCG@10 & R@50 & R@1k & latency & mAP & nDCG@10 & R@50 & R@1k & Latency \\
    \hline
    HyDE \\
    \hspace{1em}w/ 1 pseudo-doc & $48.77$ & $72.49$ & $53.20$ & $87.73$ & $8.08$ & $51.31$ & $70.37$ & $63.28$ & $87.81$ & \bm{$2.09$} \\
    \hspace{1em}w/ 1 pseudo-doc + query & $50.87$ & \bm{$75.44$} & \bm{$54.93$} & $88.76$ & \bm{$7.21$} & $50.94$ & \bm{$73.94$} & $63.80$ & $88.03$ & $2.14$ \\
    \hspace{1em}w/ 8 pseudo-doc + query & \bm{$51.64$} & $75.12$ & $54.51$ & \bm{$89.17$} & $14.15$ & \bm{$53.14$} & $73.65$ & \bm{$65.79$} & \bm{$88.67$} & $3.44$ \\
    \hline
  \end{tabular}
  \caption{\label{tab:different concatenation}
    HyDE with different concatenation of hypothetical documents and queries.
  }
\end{table*}

\begin{table*}[t]
    \centering
    \setlength{\tabcolsep}{5pt}
    \scriptsize
    \begin{tabular}{lcccccccccc}
        \hline
        \multirow{3}{*}{\textbf{Hyperparameter}}  & \multicolumn{5}{c}{\raisebox{-1mm}{\textbf{TREC DL19}}} & \multicolumn{5}{c}{\raisebox{-1mm}{\textbf{TREC DL20}}} \\
        \cmidrule(lr){2-6} \cmidrule(lr){7-11}
        & mAP & nDCG@10 & R@50 & R@1k & latency & mAP & nDCG@10 & R@50 & R@1k & Latency \\
        \hline
        Hybrid Search \\
        \hspace{1em}$\alpha$ = 0.1 & $46.0$0 & $70.87$ & $49.24$ & $88.89$ & $2.98$ & $46.54$ & $69.05$ & $63.36$ & $87.32$ & $0.90$ \\
        \hspace{1em}$\alpha$ = 0.3 & $47.14$ & \bm{$72.50$} & $51.13$ & \bm{$89.08$} & $3.20$ & \bm{$47.72$} & \bm{$69.80$} & $64.32$ & \bm{$88.04$} & \bm{$0.77$} \\
        \hspace{1em}$\alpha$ = 0.5 & \bm{$47.36$} & $72.24$ & \bm{$52.71$} & $88.09$ & $3.02$ & $47.19$ & $68.12$ & \bm{$64.90$} & $87.86$ & $0.87$ \\
        \hspace{1em}$\alpha$ = 0.7 & $47.21$ & $71.89$ & $52.40$ & $88.01$ & $3.15$ & $45.82$ & $67.30$ & $64.23$ & $87.92$ & $1.02$ \\
        \hspace{1em}$\alpha$ = 0.9 & $46.35$ & $70.67$ & $52.64$ & $88.22$ & \bm{$2.74$} & $44.02$ & $65.55$ & $63.22$ & $87.76$ & $1.20$ \\
        \hline
    \end{tabular}
    \caption{\label{tab: different alpha}Results of hybrid search with different alpha values.}
\end{table*}

\subsubsection{HyDE with Different Concatenation of Documents and Query}

Table \ref{tab:different concatenation} shows the impact of different concatenation strategies for hypothetical documents and queries using HyDE. Concatenating multiple pseudo-documents with the original query can significantly enhance retrieval performance, though at the cost of increased latency, suggesting a trade-off between retrieval effectiveness and efficiency. However, indiscriminately increasing the number of hypothetical documents does not yield significant benefits and substantially raises latency, indicating that using a single hypothetical document is sufficient.

\subsubsection{Hybrid Search with Different Weight on Sparse Retrieval}

Table \ref{tab: different alpha} presents the impact of different $\alpha$ values in hybrid search, where $\alpha$ controls the weighting between sparse retrieval and dense retrieval components. The relevance score is calculated as follows:
\vspace{-2mm}
\begin{equation}
    \small
    S_h = \alpha \cdot S_s + S_d
\end{equation}
where $S_s$, $S_d$ are the normalized relevance scores from sparse retrieval and dense retrieval respectively, and $S_h$ is the total retrieval score.

We evaluated five different $\alpha$ values to determine their influence on performance. The results indicate that an $\alpha$ value of 0.3 yields the best performance, demonstrating that appropriate adjustment of $\alpha$ can enhance retrieval effectiveness to a certain extent. Therefore, we selected $\alpha = 0.3$ for our retrieval and main experiments.
Additional implementation details are presented in Appendix \ref{Retrieval Appendix}.

\begin{table*}
  \centering
  \small
  \setlength{\tabcolsep}{5pt}
  \begin{tabular}{lccccccc}
    \hline
    \multirow{3}{*}{\textbf{Method}} & \multicolumn{7}{c}{\raisebox{-1mm}{\textbf{MS MARCO Passage ranking}}} \\
    \cmidrule(lr){2-8}
    & Base Model & \# Params & MRR@1 & MRR@10 & MRR@1k & Hit Rate@10 & Latency \\
    \hline
    \textit{w/o Reranking} \\
    Random Ordering & - & - & $0.011$ & $0.027$ & $0.068$ & $0.092$ & - \\
    BM25 & - & - & $6.52$ & $11.65$ & $12.59$ & $24.63$ & - \\
    \hline
    \textit{DLM Reranking} \\
    monoT5 & T5-base & 220M & $21.62$ & $31.78$ & $32.40$ & $54.07$ & \bm{$4.5$} \\
    monoBERT & BERT-large & 340M & $21.65$ & $31.69$ & $32.35$ & $53.38$ & $15.8$ \\
    RankLLaMA & Llama-2-7b & 7B & \bm{$22.08$} & \bm{$32.35$} & \bm{$32.97$} & \bm{$54.53$} & $82.4$ \\
    \hline
    \textit{TILDE Reranking} \\
    TILDEv2 & BERT-base & 110M & $18.57$ & $27.83$ & $28.60$ & $49.07$ & \bm{$0.02$} \\
    \hline
  \end{tabular}
  \caption{
    Results of different reranking methods on the dev set of the MS MARCO Passage ranking dataset. For each query, the top-1000 candidate passages retrieved by BM25 are reranked. Latency is measured in seconds per query.
  }
  \label{tab:reranking_table}
  \vspace{-4mm}
\end{table*}
 
\subsection{Reranking Methods}
 
After the initial retrieval, a reranking phase is employed to enhance the relevance of the retrieved documents, ensuring that the most pertinent information appears at the top of the list. 
This phase uses more precise and time-intensive methods to reorder documents effectively, increasing the similarity between the query and the top-ranked documents.

We consider two approaches in our reranking module: \textbf{DLM Reranking}, which utilizes classification, and \textbf{TILDE Reranking}, which focuses on query likelihoods. These approaches prioritize performance and efficiency, respectively.

\begin{itemize}[left=0pt]
\setlength{\parsep}{0pt}
\setlength{\parskip}{0pt}
\item \textbf{DLM Reranking:}\quad 
This method leverages deep language models (DLMs)~\cite{nogueira2019multi, nogueira2020document, ma2023fine} for reranking. These models are fine-tuned to classify document relevancy to a query as ``true'' or ``false''. During fine-tuning, the model is trained with concatenated query and document inputs, labeled by relevancy. At inference, documents are ranked based on the probability of the ``true'' token.

\item \textbf{TILDE Reranking:}\quad 
TILDE~\cite{zhuang2021tilde, zhuang2021fast} calculates the likelihood of each query term independently by predicting token probabilities across the model’s vocabulary. Documents are scored by summing the pre-calculated log probabilities of query tokens, allowing for rapid reranking at inference. TILDEv2 improves this by indexing only document-present tokens, using NCE loss, and expanding documents, thus enhancing efficiency and reducing index size.
\end{itemize}
\ 

Our experiments were conducted on the MS MARCO Passage ranking dataset~\cite{bajaj2016ms}, a large-scale dataset for machine reading comprehension. 
We follow and make modifications to the implementation provided by PyGaggle~\cite{nogueira2020document} and TILDE~\cite{zhuang2021tilde}, using the models monoT5, monoBERT, RankLLaMA and TILDEv2.
Reranking results are shown in Table~\ref{tab:reranking_table}. 
We recommend \underline{\textbf{monoT5}} as a comprehensive method balancing performance and efficiency. 
\textbf{RankLLaMA} is suitable for achieving the best performance, while \textbf{TILDEv2} is ideal for the quickest experience on a fixed collection.
Details on the experimental setup and results are presented in Appendix \ref{Reranking Appendix}.

\subsection{Document Repacking}
 
The performance of subsequent processes, such as LLM response generation, may be affected by the order documents are provided. 
To address this issue, we incorporate a compact repacking module into the workflow after reranking, featuring three repacking methods: \textbf{``forward''}, \textbf{``reverse''} and \textbf{``sides''}. 
The ``forward'' method repacks documents by descending relevancy scores from the reranking phase, whereas the ``reverse'' arranges them in ascending order. 
Inspired by~\citet{liu2024lost}, concluding that optimal performance is achieved when relevant information is placed at the head or tail of the input, we also include a ``sides'' option.

Since the repacking method primarily affects subsequent modules, we select the best repacking method in Section \ref{section4} by testing it in combination with other modules.
In this section, we choose the \underline{\textbf{``sides''}} method as the default repacking method.

\begin{table*}[t]
  \centering
      \small
      \setlength{\tabcolsep}{5pt}
  \begin{tabular}{lcccccccccccc}
    \hline
    \multirow{2}{*}{\raisebox{-1mm}{\textbf{Method}}} & \multicolumn{2}{c}{\raisebox{-1mm}{\textbf{NQ}}} & \multicolumn{2}{c}{\raisebox{-1mm}{\textbf{TQA}}} & \multicolumn{2}{c}{\raisebox{-1mm}{\textbf{HotPotQA}}} & \multirow{2}{*} {\raisebox{-1mm}{\textbf{Avg.}}}  & \multirow{2}{*}{\raisebox{-1mm}{\textbf{Avg. Token}}} \\ 
    \cmidrule(lr){2-3} \cmidrule(lr){4-5} \cmidrule(lr){6-7}
    &F1 & \#token & F1 & \#token &F1 & \#token
    \\ \hline
    
    \textit{w/o Summarization}   \\
    Origin Prompt  &$27.07$ & $124$ & $33.61$ & $152$ & $33.92$& $141$ & $31.53$ & $139$ \\
    \hline
    \textit{Extractive Method}                                                 \\
    BM25               & $27.97$ & $40$ & $32.44$ & $59$ & $28.00$ & $63$ & $29.47$   & $54$ \\
    Contriever         &  $23.62$  & $42$ & $33.79$ & $65$ & $23.64$ & $60$ & $27.02$ &  $56$\\
    Recomp (extractive) & $27.84$ & $34$ & $35.32$ & $60$ & $29.46$ & $58$ & $30.87$  & $51$ \\
    \hline
    \textit{Abstractive Method} \\
    SelectiveContext & $25.05$ & $65$ & $34.25$ & $70$ & \bm{$34.43$}  & $66$ & $31.24$ &$67$   \\
    LongLLMlingua   &$21.32$  & $51$ &$32.81$ & $56$ & $30.79$ & $57$ & $28.29$ & $55$ \\
    Recomp (abstractive) & \bm{$33.68$}  & $59$ & \bm{$35.87$} &$61$ & $29.01$& $57$ & \bm{$32.85$} & $59$ \\
    
    \hline
  \end{tabular}
  \caption{
    Comparison between different summarization methods.   }
    \label{tab:compression_table}
    \vspace{-3mm}
\end{table*}

\subsection{Summarization}
 
Retrieval results may contain redundant or unnecessary information, potentially preventing LLMs from generating accurate responses. 
Additionally, long prompts can slow down the inference process. 
Therefore, efficient methods to summarize retrieved documents are crucial in the RAG pipeline.

Summarization tasks can be \textbf{extractive} or \textbf{abstractive}. 
Extractive methods segment text into sentences, then score and rank them based on importance. 
Abstractive compressors synthesize information from multiple documents to rephrase and generate a cohesive summary. 
These tasks can be query-based or non-query-based.
In this paper, as RAG retrieves information relevant to queries, we focus exclusively on query-based methods.
 
\begin{itemize}[left=0pt] 
\setlength{\parsep}{0pt}
\setlength{\parskip}{0pt}
\item \textbf{Recomp:} \quad
Recomp~\cite{xu2023recomp} has extractive and abstractive compressors. 
The extractive compressor selects useful sentences, while the abstractive compressor synthesizes information from multiple documents. 
\item \textbf{LongLLMLingua:} \quad
LongLLMLingua~\cite{jiang2023longllmlingua} improves LLMLingua by focusing on key information related to the query. 
\item \textbf{Selective Context} \quad
Selective Context enhances LLM efficiency by identifying and removing redundant information in the input context. It evaluates the informativeness of lexical units using self-information computed by a base causal language model. This method is non-query-based, allowing a comparison between query-based and non-query-based approaches.

\end{itemize}

We evaluate these methods on three benchmark datasets: NQ, TriviaQA, and HotpotQA. 
Comparative results of different summarization methods are shown in Table \ref{tab:compression_table}. 
We recommend \underline{\textbf{Recomp}} for its outstanding performance. 
LongLLMLingua does not perform well but demonstrates better generalization capabilities as it was not trained on these experimental datasets. Therefore, we consider it as an alternative method.
Additional implementation details and discussions on non-query-based methods are provided in Appendix \ref{Summarization Appendix}.

\subsection{Generator Fine-tuning}
\label{sec:fine-tune}
 
In this section, we focus on fine-tuning the generator while leaving retriever fine-tuning for future exploration. 
We aim to investigate the impact of fine-tuning, particularly the influence of relevant or irrelevant contexts on the generator's performance.

Formally, we denote \( x \) as the query fed into the RAG system, and \(\mathcal{D}\) as the contexts for this input. The fine-tuning loss of the generator is the negative log-likelihood of the ground-truth output \( y \).
%

To explore the impact of fine-tuning, especially relevant and irrelevant contexts, we define \( d_{gold} \) as a context relevant to the query, and \( d_{random} \) as a randomly retrieved context. We train the model by varying the composition of \(\mathcal{D}\) as follows:
{
\begin{itemize} [left=0pt]
\item \( D_{g} \): The augmented context consists of query-relevant documents, denoted as \(D_{g} = \{d_{gold}\}\).
\item \( D_{r} \): The context contains one randomly sampled document, denoted as \(D_{r} = \{d_{random}\}\).
\item \( D_{gr} \): The augmented context comprises a relevant document and a randomly-selected one, denoted as \(D_{gr} = \{d_{gold}, d_{random}\}\).
\item \( D_{gg} \): The augmented context consists of two copies of a query-relevant document, denoted as \(D_{gg} = \{d_{gold}, d_{gold}\}\).
\end{itemize}
}

We denote the base LM generator not fine-tuned as $M_{b}$ , and the model fine-tuned under the corresponding $\mathcal{D}$ as $M_{g}$, $M_{r}$, $M_{gr}$, $M_{gg}$.
We fine-tuned our model on several QA and reading 
\begin{wrapfigure}{r}{0.5\textwidth}
 \centering
  \setlength{\abovecaptionskip}{1mm}
 \includegraphics[width=0.5\textwidth]{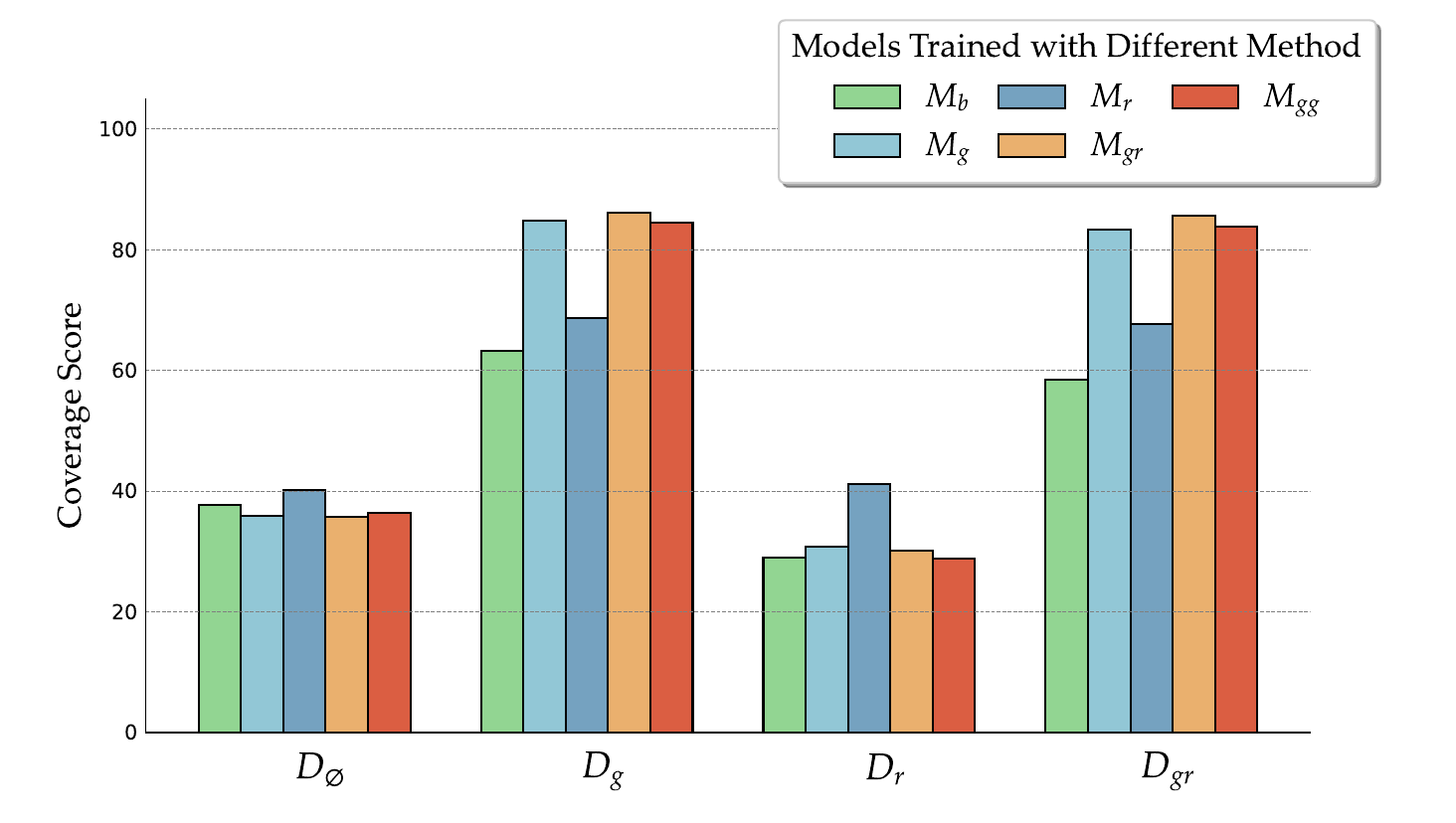}
  \caption{\label{fig: finetune}Results of generator fine-tuning. 
  }
\end{wrapfigure}
comprehension datasets. Ground-truth coverage is used as our evaluation metric since QA task answers are relatively short.
We select Llama-2-7B~\cite{Touvron2023Llama2O} as the base model. 
Similar to training, we evaluate all trained models on validation sets with \( D_{g} \), \( D_{r} \), \( D_{gr} \), and \( D_\varnothing \), where \( D_\varnothing \) indicates inference without retrieval.
Figure \ref{fig: finetune} presents our main results. Models trained with a mix of relevant and random documents (\(M_{gr}\)) perform best when provided with either gold or mixed contexts. This suggests that mixing relevant and random contexts during training can enhance the generator's robustness to irrelevant information while ensuring effective utilization of relevant contexts. 
Therefore, we identify the practice of augmenting with a few \textbf{\underline{relevant and randomly-selected} \underline{documents during training}} as the best approach.
Detailed dataset information, hyperparameters and experimental results can be found in Appendix \ref{Generator Fine-tuning Appendix}.

\section{Searching for Best RAG Practices}
\label{section4}
 
In the following section, we investigate the optimal practices for implementing RAG. 
To begin with, we used the default practice identified in Section \ref{workflow} for each module. Following the workflow depicted in Figure \ref{fig:workflow}, we sequentially optimized individual modules and selected the most effective option among alternatives. 
This iterative process continued until we determined the best method for implementing the final summarization module.
Based on Section \ref{sec:fine-tune}, we used the Llama2-7B-Chat model fine-tuned where each query was augmented by a few random-selected and relevant documents as the generator. 
We used Milvus to build a vector database that includes $10$ million text of English Wikipedia and $4$ million text of medical data.
We also investigated the impact of removing the Query Classification, Reranking, and Summarization modules to assess their contributions. 
 
\subsection{Comprehensive Evaluation}
 
We conducted extensive experiments across various NLP tasks and datasets to assess the performance of RAG systems. Specifically:
(I) \textbf{Commonsense Reasoning}; 
(II) \textbf{Fact Checking}; 
(III) \textbf{Open-Domain QA}; 
(IV) \textbf{MultiHop QA}; 
(V) \textbf{Medical QA}.
For further details on the tasks and their corresponding datasets, please refer to Appendix \ref{Evaluation Appendix}.
Furthermore, we evaluated the \textbf{RAG capabilities} on subsets extracted from these datasets, employing the metrics recommended in RAGAs~\cite{Shahul2023RAGAsAE}, including Faithfulness, Context Relevancy, Answer Relevancy, and Answer Correctness. Additionally, we measured Retrieval Similarity by computing the cosine similarity between retrieved documents and gold documents.


We used accuracy as the evaluation metric for the tasks of Commonsense Reasoning, Fact Checking, and Medical QA. For Open-Domain QA and Multihop QA, we employed token-level F1 score and Exact Match (EM) score.
The final RAG score was calculated by averaging the aforementioned five RAG capabilities.
We followed~\citet{Trivedi_Balasubramanian_Khot_Sabharwal_2022}  and sub-sampled up to $500$ examples from each dataset.


\begin{table*}[t]
    \centering
    \scriptsize
    \setlength{\tabcolsep}{3pt}
    \renewcommand{\arraystretch}{1.1}
\begin{tabular}{lccccccccccc}

\toprule
\multirow{2}{*}{\bf Method} & \textbf{Commonsense} & \textbf{Fact Check} & \multicolumn{2}{c}{\textbf{ODQA}} & \multicolumn{2}{c}{\textbf{Multihop}} & \textbf{Medical} & \textbf{RAG} & \multicolumn{3}{c}{\textbf{Avg.}} \\
\cmidrule(lr){2-2}
\cmidrule(lr){3-3}
\cmidrule(lr){4-5}
\cmidrule(lr){6-7}
\cmidrule(lr){8-8}
\cmidrule(lr){9-9}
\cmidrule(lr){10-12}
 & Acc & Acc & EM & F1 & EM & F1 & Acc & Score & Score & F1 & Latency \\

\hline
\multicolumn{12}{c}{\framebox{\textit{classification module}} , Hybrid with HyDE, monoT5, sides, Recomp} \rule{0pt}{11.5pt} \\
w/o classification  & 0.719 & 0.505 & 0.391 & \bf 0.450 & \bf 0.212 & 0.255 & \bf 0.528 & 0.540 & 0.465 & \bf 0.353 & 16.58 \rule{0pt}{8pt} \\
\underline{+ \textbf{classification}} & \bf 0.727 & \bf 0.595 & \bf 0.393 & \bf 0.450 & 0.207 & \bf 0.257 & 0.460 & \bf 0.580 & \bf 0.478 & \bf 0.353  & \bf 11.71 \\
\noalign{\vskip 1.5pt}
\hdashline 
\multicolumn{12}{c}{with classification, \framebox{\textit{ retrieval module}} , monoT5, sides, Recomp} \rule{0pt}{11.5pt} \\
+ HyDE & 0.718 & \bf 0.595 & 0.320 & 0.373 & 0.170 & 0.213 & 0.400 & 0.545 & 0.443 & 0.293 & 11.58 \rule{0pt}{8pt} \\
+ Original & 0.721 & 0.585 & 0.300 & 0.350 & 0.153 & 0.197 & 0.390 & 0.486 & 0.428 & 0.273 & \bf 1.44 \\
+ Hybrid & 0.718 & \bf 0.595 & 0.347 & 0.397 & 0.190 & 0.240 & \bf 0.750 & 0.498 & 0.477 & 0.318 & 1.45 \\
\underline{+ \textbf{Hybrid with HyDE}} & \bf 0.727 & \bf 0.595 & \bf 0.393 & \bf 0.450 & \bf 0.207 & \bf 0.257 & 0.460 & \bf 0.580 & \bf 0.478 & \bf 0.353  & 11.71 \\
\noalign{\vskip 2.5pt}
\hdashline
\multicolumn{12}{c}{with classification, Hybrid with HyDE, \framebox{\textit{reranking module}} , sides, Recomp} \rule{0pt}{11.5pt} \\
w/o reranking & 0.720 & 0.591 & 0.365 & 0.429 & 0.211 & \bf 0.260 & \bf 0.512 & 0.530 & 0.470 & 0.334 & \bf 10.31 \\
\underline{+ \textbf{monoT5}} & \bf 0.727 & 0.595 & 0.393 & 0.450 & 0.207 & 0.257 & 0.460 & \bf 0.580 & \bf 0.478 & 0.353  & 11.71 \\
+ monoBERT & 0.723 & 0.593 & 0.383 & 0.443 & \bf 0.217 & 0.259 & 0.482 & 0.551 & 0.475 & 0.351 & 11.65 \\
+ RankLLaMA & 0.723 & \bf 0.597 & 0.382 & 0.443 & 0.197 & 0.240 & 0.454 & 0.558 & 0.470 & 0.342 & 13.51 \\
+ TILDEv2 & 0.725 & 0.588 & \bf 0.394 & \bf 0.456 & 0.209 & 0.255 & 0.486 & 0.536 & 0.476 & \bf 0.355 & 11.26 \\
\noalign{\vskip 1pt}
\hdashline
\multicolumn{12}{c}{with classification, Hybrid with HyDE, monoT5, \framebox{\textit{repacking module}} , Recomp} \rule{0pt}{11.5pt} \\
+ sides & 0.727 & 0.595 & \bf 0.393 & \bf 0.450 & 0.207 & 0.257 & 0.460 & \bf 0.580 & 0.478 & 0.353  & 11.71 \\
+ forward & 0.722 & \bf 0.599 & 0.379 & 0.437 & 0.215 & 0.260 & 0.472 & 0.542 & 0.474 & 0.349 & \bf 11.68 \\
\underline{+ \textbf{reverse}} & \bf 0.728 & 0.592 & 0.387 & 0.445 & \bf 0.219 & \bf 0.263 & \bf 0.532 & 0.560 & \bf 0.483 & \bf 0.354 & 11.70 \\
\noalign{\vskip 1.5pt}
\hdashline
\multicolumn{12}{c}{with classification, Hybrid with HyDE, monoT5, reverse, \framebox{\textit{summarization module}}} \rule{0pt}{11.5pt} \\
w/o summarization & \bf 0.729 & 0.591 & \bf 0.402  & \bf 0.457 & 0.205 & 0.252 & 0.528 & 0.533 & 0.480 & \bf 0.355 & \bf 10.97 \\
\underline{+ \textbf{Recomp}} & 0.728 & \bf 0.592 & 0.387 & 0.445 & \bf 0.219 & \bf 0.263 & \bf 0.532 & \bf 0.560 & \bf 0.483 & 0.354 & 11.70 \\
+ LongLLMLingua & 0.713 & 0.581 & 0.362  & 0.423 & 0.199 & 0.245 & 0.530 & 0.539 & 0.466 & 0.334 & 16.17 \\

\bottomrule

\end{tabular}
 
\caption{
     Results of the search for optimal RAG practices. Modules enclosed in a \framebox{boxed module} are under investigation to determine the best method. The \underline{\textbf{underlined method}} represents the selected implementation. The ``Avg'' (average score) is calculated based on the Acc, EM, and RAG scores for all tasks, while the average latency is measured in seconds per query. The best scores are highlighted in \textbf{bold}.}
    \label{tab:final_tabel}
     
\end{table*}

\subsection{Results and Analysis}

Based on the experimental results presented in Table \ref{tab:final_tabel}, the following key insights emerge:

\begin{itemize}[left=0pt] 
\setlength{\parsep}{0pt}
\setlength{\parskip}{0pt}
\item \textbf{Query Classification Module:} 
This module is referenced and contributes to both effectiveness and efficiency, leading to an average improvement in the overall score from $0.428$ to $0.443$ and a reduction in latency time from $16.41$ to $11.58$ seconds per query.

\item \textbf{Retrieval Module:} While the ``Hybrid with HyDE'' method attained the highest RAG score of $0.58$, it does so at a considerable computational cost with $11.71$ second per query. Consequently, the ``Hybrid'' or ``Original'' methods are recommended, as they reduce latency while maintaining comparable performance.

\item \textbf{Reranking Module:} 
The absence of a reranking module led to a noticeable drop in performance, highlighting its necessity. 
MonoT5 achieved the highest average score, affirming its efficacy in augmenting the relevance of retrieved documents. 
This indicates the critical role of reranking in enhancing the quality of generated responses.

\item \textbf{Repacking Module:}  
The Reverse configuration exhibited superior performance, achieving an RAG score of $0.560$. This indicates that positioning more relevant context closer to the query leads to optimal outcomes.

\item \textbf{Summarization Module:} 
Recomp demonstrated superior performance, although achieving comparable results with lower latency was possible by removing the summarization module. Nevertheless, Recomp remains the preferred choice due to its capability to address the generator's maximum length constraints. In time-sensitive applications, removing summarization could effectively reduce response time.

\end{itemize}

The experimental results demonstrate that each module contributes uniquely to the overall performance of the RAG system. The query classification module enhances accuracy and reduces latency, while the retrieval and reranking modules significantly improve the system's ability to handle diverse queries. The repacking and summarization modules further refine the system's output, ensuring high-quality responses across different tasks.

\section{Discussion}

\subsection{Best Practices for Implementing RAG}
According to our experimental findings, we suggest two distinct recipes or practices for implementing RAG systems, each customized to address specific requirements: one focusing on maximizing performance, and the other on striking a balance between efficiency and efficacy.

\noindent \textbf{Best Performance Practice:}
To achieve the highest performance, it is recommended to incorporate query classification module, use the ``Hybrid with HyDE'' method for retrieval, employ monoT5 for reranking, opt for Reverse for repacking, and leverage Recomp for summarization. 
This configuration yielded the highest average score of $0.483$, albeit with a computationally-intensive process.

\noindent \textbf{Balanced Efficiency Practice:}
In order to achieve a balance between performance and efficiency, it is recommended to incorporate the query classification module, implement the Hybrid method for retrieval, use TILDEv2 for reranking, opt for Reverse for repacking, and employ Recomp for summarization. 
Given that the retrieval module accounts for the majority of processing time in the system, transitioning to the Hybrid method while keeping other modules unchanged can substantially reduce latency while preserving a comparable performance.


\begin{figure*}[t]
    \centering
    \includegraphics[width=0.9\linewidth]{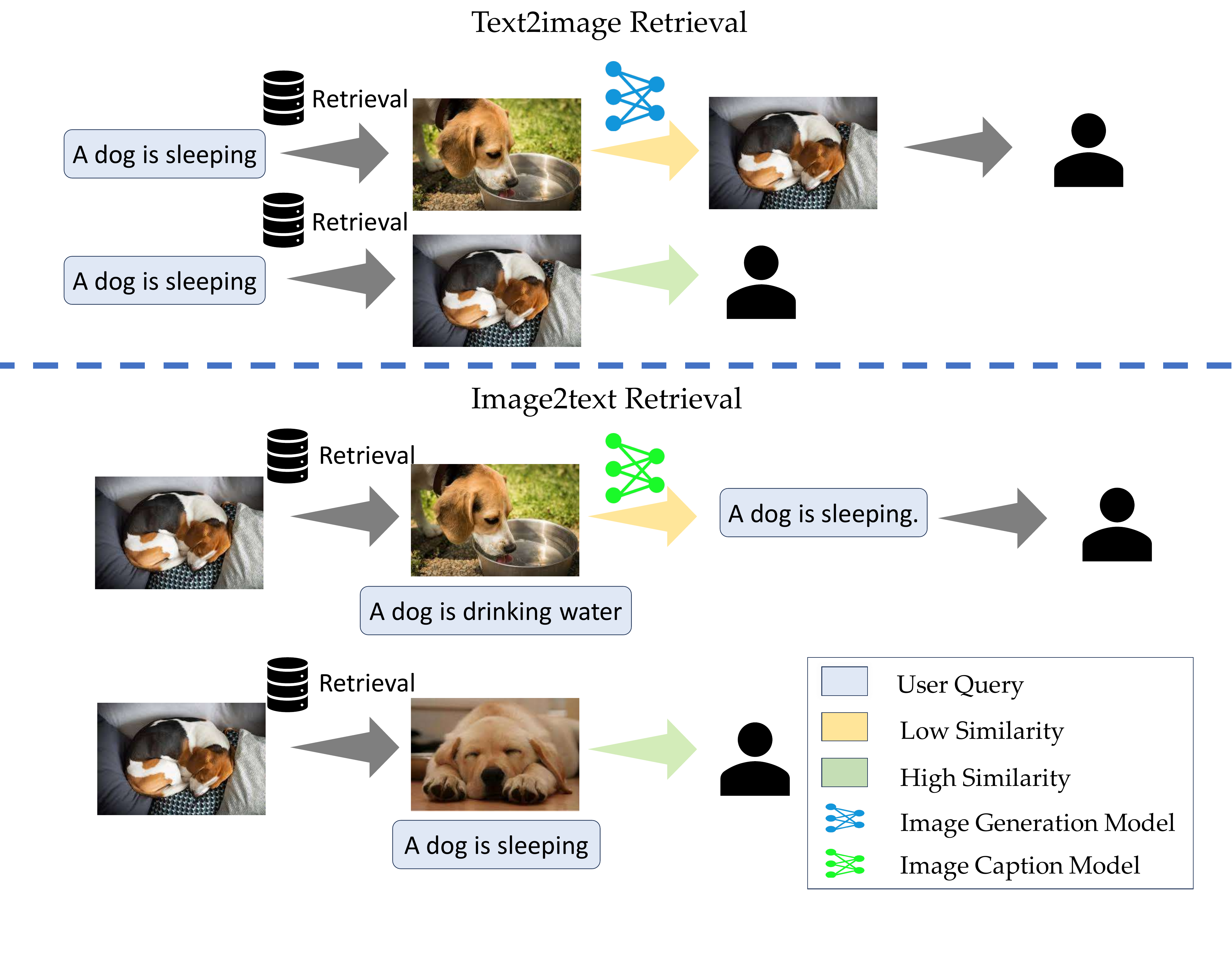}
    \vspace{-3mm}
    \caption{\label{fig:mutimodal-retriever} Workflow of multimodal retrieval. The upper section illustrates the text-to-image retrieval process. Initially, a text query is used to find images in the database with the highest similarity. If a high similarity is found, the image is returned directly. If not, an image generation model is employed to create and return an appropriate image.
The lower section demonstrates the image-to-text retrieval process. Here, a user-provided image is matched with images in the database to find the highest similarity. If a high similarity is identified, the pre-stored caption of the matching image is returned. Otherwise, an image captioning model generates and returns a new caption.
    \label{mutimodal:workflow}}
\end{figure*}

\subsection{Multimodal Extension}

We have extended RAG to multimodal applications. Specifically, we have incorporated text2image and image2text retrieval capabilities into the system with a substantial collection of paired image and textual descriptions as a retrieval source.
As depicted in Figure \ref{mutimodal:workflow}, the text2image capability speeds up the image generation process when a user query aligns well with the textual descriptions of stored images (i.e., ``retrieval as generation'' strategy), while the image2text functionality comes into play when a user provides an image and engages in conversation about the input image.
These multimodal RAG capabilities offer the following advantages:


\begin{itemize}[left=0pt] 
\setlength{\parsep}{0pt}
\setlength{\parskip}{0pt}
\item \textbf{Groundedness}: Retrieval methods provide information from verified multimodal materials, thereby ensuring authenticity and specificity. In contrast, on-the-fly generation relies on models to generate new content, which can occasionally result in factual errors or inaccuracies.

\item  \textbf{Efficiency}: Retrieval methods are typically more efficient, especially when the answer already exists in stored materials. Conversely, generation methods may require more computational resources to produce new content, particularly for images or lengthy texts.

\item  \textbf{Maintainability}: Generation models often necessitate careful fine-tuning to tailor them for new applications. In contrast, retrieval-based methods can be improved to address new demands by simply enlarging the size and enhancing the quality of retrieval sources.
\end{itemize}

We plan to broaden the application of this strategy to include other modalities, such as video and speech, while also exploring efficient and effective cross-modal retrieval techniques.



\section{Conclusion}

In this study, we aim to identify optimal practices for implementing retrieval-augmented generation in order to improve the quality and reliability of content produced by large language models. 
We systematically assessed a range of potential solutions for each module within the RAG framework and recommended the most effective approach for each module. 
Furthermore, we introduced a comprehensive evaluation benchmark for RAG systems and conducted extensive experiments to determine the best practices among various alternatives.
Our findings not only contribute to a deeper understanding of retrieval-augmented generation systems but also establish a foundation for future research.


\section*{Limitations}

We have evaluated the impact of various methods for fine-tuning LLM generators. 
Previous studies have demonstrated the feasibility of training both the retriever and generator jointly. 
We would like to explore this possibility in the future. 
In this study, we embraced the principle of modular design to simplify the search for optimal RAG implementations, thereby reducing complexity. 
Due to the daunting costs associated with constructing vector databases and conducting experiments, our evaluation was limited to investigating the effectiveness and influence of representative chunking techniques within the chunking module. It would be intriguing to further explore the impact of different chunking techniques on the entire RAG systems.
While we have discussed the application of RAG in the domain of NLP and extended its scope to image generation, an enticing avenue for future exploration would involve expanding this research to other modalities such as speech and video.



\section*{Acknowledgments}

The authors would like to thank the anonymous reviewers for their valuable comments. This work was supported by National Natural Science Foundation of China (No. $62076068$).


\bibliographystyle{unsrtnat}

\bibliography{neurips_2024}

\clearpage
\appendix

\section{Experimental Details}
\label{sec:appendix}
In this section, we provide detailed experimental settings for each module, covering dataset specifics,
training parameters, and any additional experimental results.

\subsection{Query Classification}
\label{Query Classification Appendix}
\noindent \textbf{Datasets} \quad   We utilized a subset of the Databricks-Dolly-15K~\cite{DatabricksBlog2023DollyV2} and generated additional data using GPT-4.The prompt template for generating questions is shown in Table \ref{tab:task generate}.

\noindent \textbf{Implementation Details}\quad   We choose  BERT-base-multilingual-cased as our classifier, with a batch size of 16 and a learning rate of 1e-5. The evaluation of results is showcased in Table \ref{tab:task cls}.


\subsection{Experimental Details of Retrieval Methods}
\label{Retrieval Appendix}

Implementation details of the comparative experiments of different retrieval methods are as below:

\noindent\textbf{Datasets}
\quad We use the TREC DL 2019~\cite{Craswell2020OverviewOT} and 2020~\cite{Craswell2021OverviewOT} passage ranking datasets to evaluate the performance of different retrieval methods.\\
\noindent\textbf{Metrics} 
\quad Widely-used evaluation metrics for retrieval include mAP, nDCG@10, R@50 and R@1k. Both mAP and nDCG@10 are order-aware metrics that take the ranking of search results into account. In contrast, R@k is an order-unaware metric. We also report the average latency incurred by each method per query. \\
\noindent\textbf{Implementation Details} 
\quad For sparse retrieval, we use the BM25 algorithm, which relies on the TF-IDF algorithm. For dense retrieval, we employ Contriever as our unsupervised contrastive text encoder. Based on our evaluation of embedding models, we implement our supervised dense retrieval using LLM-Embedder.
We use the default implementation of BM25 and Contriever from Pyserini~\cite{lin2021pyserini}.
The BM25 index is constructed using Lucene on MS MARCO collections, while the dense vector index is generated with Faiss employing Flat configuration on the same dataset.
For query rewriting, we prompt Zephyr-7b-alpha\footnote{\url{https://huggingface.co/HuggingFaceH4/zephyr-7b-alpha}}, a model trained to act as a helpful assistant, to rewrite the original query. 
For query decomposition, we employ GPT-3.5-turbo-0125 to break down the original query into multiple sub-queries.
We closely follow the implementation from HyDE~\cite{gao2022precise}, utilizing the more advanced instruction-following language model, GPT-3.5-turbo-instruct, to generate hypothetical answers. The model infers with a default temperature of 0.7, sampling up to a maximum of 512 tokens. Retrieval experiments and evaluation are conducted using the Pyserini toolkit.

\subsection{Experimental Details of Reranking Methods}
\label{Reranking Appendix}

\noindent\textbf{Datasets}
\quad Our experiments utilize the MS MARCO Passage ranking dataset, a substantial corpus designed for machine reading comprehension tasks. This dataset comprises over 8.8 million passages and 1 million queries. The training set contains approximately 398M tuples of queries paired with corresponding positive and negative passages, while the development set comprises 6,980 queries, paired with their BM25 retrieval results, and preserves the top-1000 ranked candidate passages for each query. We evaluate the effectiveness of the methods on the development set, as the test set is not publicly available.

\noindent\textbf{Metrics} 
\quad The evaluation metrics MRR@1, MRR@10, MRR@1k and Hit Rate@10 are used. MRR@10 is the official metric proposed by MS MARCO.

\noindent\textbf{Implementation Details} 
\quad We follow and make modifications to the implementation provided by PyGaggle~\cite{nogueira2020document} and TILDE~\cite{zhuang2021tilde}. For DLM-based reranking, we use monoT5~\cite{nogueira2020document} based on T5-base, monoBERT~\cite{nogueira2019multi} based on BERT-large and RankLLaMA~\cite{ma2023fine} based on Llama-2-7b. For TILDE reranking, we use TILDEv2~\cite{zhuang2021fast} based on BERT-base.

Typically, 50 documents are retrieved as input for the reranking module. The documents remaining after the reranking and repacking phase can be further concentrated by assigning a top-k value or a relevancy score threshold. 

\noindent \textbf{Result Analysis}
\quad Reranking results are shown in Table~\ref{tab:reranking_table}. 
We compare our results with a randomly shuffled ordering and the BM25 retrieval baseline. 
All reranking methods demonstrate a notable increase in performance across all metrics. 
Approximately equal performance is achieved by monoT5 and monoBERT, and RankLLaMA performs best, each ascending in latency. 
TILDEv2 is the fastest, taking approximately 10 to 20 milliseconds per query at the cost of performance. 
Additionally, TILDEv2 requires that the passages reranked be identically included in the previously indexed collection. 
Preprocessing must be redone at inference for new unseen passages, negating the efficiency advantages.

\subsection{Experimental Details of Summarization Methods}
\label{Summarization Appendix}

\noindent \textbf{Selective Context} \quad
Selective Context enhances LLM efficiency by identifying and removing redundant information in the input context. It evaluates the informativeness of lexical units using self-information computed by a base causal language model. This method is non-query-based, allowing a comparison between query-based and non-query-based approaches.

\noindent \textbf{Datasets} \quad
We evaluated these methods on three datasets: Natural Questions (NQ)~\cite{Kwiatkowski2019NaturalQA}, TriviaQA~\cite{Joshi2017TriviaQAAL}, and HotpotQA~\cite{yang2018hotpotqa}. 

\noindent \textbf{Metrics} \quad
Evaluation metrics include the F1 score and the number of tokens changed after summarization to measure conciseness.

\noindent \textbf{Implementation Details} \quad
For all methods, we use Llama3-8B-Instruct as the generator model and set a summarization ratio of 0.4. For extractive methods, importance scores determine the sentences retained. For abstractive methods, we control the maximum generation length using the summarization ratio to align with extractive methods.
Experiments are conducted on the NQ test set, TriviaQA test set, and HotpotQA development set.

\begin{table*}[t]
\small
    \centering
    
    \setlength{\tabcolsep}{9pt}
    \renewcommand{\arraystretch}{1.3}
    \begin{tabular}%
    {c|lccccc}
    \hline
        \textbf{Context} & \textbf{Model} & \textbf{NQ} & \textbf{TriviaQA}  & \textbf{HotpotQA} & \textbf{ASQA} & \textbf{Avg.}\\
        \hline
        \multirow{5}{*}{\resizebox{0.66cm}{!}{$D_\varnothing$}}&$M_{b}$ & $29.78$&$60.44$ & $23.73$& $37.89$&$37.96$ \\
         &$M_g$ & $26.2$3& $58.26$&$26.67$ &$32.30$ & $35.87$\\
         &$M_{r}$&$31.10$&$61.37$&$28.40$&$39.96$&$40.21$ \\       
         &$M_{gr}$&$25.92$&$57.62$&$26.43$&$32.99$&$35.70$\\
         &$M_{gg}$&$26.69$&$58.07$&$27.04$&$33.75$&$36.39$\\
        \hline
        \multirow{5}{*}{\resizebox{0.6cm}{!}{$D_g$}}& $M_{b}$&$44.78$&$79.90$&$56.72$&$71.64$&$63.26$\\
         &$M_g$&$85.72$&$88.16$&$79.82$&$85.51$&$84.80$\\
         &$M_{r}$&$60.98$&$80.20$&$65.73$&$67.49$&$68.60$ \\         &$M_{gr}$&$87.60$&$87.94$&$\boldsymbol{81.07}$&$87.58$&$\boldsymbol{86.05}$\\
         &$M_{gg}$&$86.72$&$\boldsymbol{88.35}$&$79.59$&$83.44$&$84.53$\\
        \hline
        \multirow{5}{*}{\resizebox{0.6cm}{!}{$D_r$}}&$M_{b}$&$16.49$&$50.03$&$21.57$&$28.79$&$29.22$\\
         &$M_g$&$22.15$&$46.98$&$24.36$&$29.40$&$30.72$\\
         &$M_r$&$36.92$&$58.42$&$29.64$&$39.54$&$41.13$\\
        &$M_{gr}$&$23.63$&$45.01$&$24.17$&$27.95$&$30.19$\\
         &$M_{gg}$&$21.08$&$43.83$&$23.23$&$27.33$&$28.87$\\
        \hline
        \multirow{5}{*}{\resizebox{0.8cm}{!}{$D_{gr}$}}&$M_{b}$&$34.65$&$81.27$&$52.75$&$65.42$&$58.52$\\
         &$M_g$&$85.00$&$87.33$&$78.18$&$83.02$&$83.38$\\
         &$M_r$&$60.28$&$79.32$&$63.82$&$67.29$&$67.68$\\         &$M_{gr}$&$\boldsymbol{87.63}$&$87.14$&$79.95$&$\boldsymbol{87.78}$&$85.63$\\
         &$M_{gg}$&$86.31$&$86.90$&$78.10$&$83.85$&$83.79$\\
        \hline
    \end{tabular}
    \caption{Results of the model augmented with different contexts on various QA datasets. }
     \label{tab:ft result}
\end{table*}

\subsection{Experimental Details of Generator Fine-tuning}
\label{Generator Fine-tuning Appendix}
\noindent\textbf{Datasets}
\quad We fine-tune our model on several question answering(QA) and reading comprehension datasets, including ASQA~\cite{Stelmakh2022ASQAFQ}, HotpotQA~\cite{yang2018hotpotqa}, NarrativeQA~\cite{kovcisky2018narrativeqa}, NQ~\cite{Kwiatkowski2019NaturalQA}, SQuAD~\cite{rajpurkar2016squad}, TriviaQA~\cite{Joshi2017TriviaQAAL}, TruthfulQA~\cite{lin2021truthfulqa}. 
We use their train splits (for those containing significantly more data
entries than others, we conducted a random sample). For evaluation, ASQA~\cite{Stelmakh2022ASQAFQ}, HotpotQA~\cite{yang2018hotpotqa}, NQ~\cite{Kwiatkowski2019NaturalQA}, TriviaQA~\cite{Joshi2017TriviaQAAL} are used. We evaluate our model on their validation splits or manually split a
\begin{wraptable}{r}{0.5\textwidth}
    \centering
    \setlength{\tabcolsep}{10pt}
    \small
        \begin{tabular}{l|r|r}
            \hline
            \textbf{Dataset} & \#\textbf{Train} & \#\textbf{Eval} \\
            \hline
            ASQA   & $2,090$   &  $483$    \\
            HotpotQA   & $15,000$   & $7,405$   \\
            TriviaQA & $9,000$ & $6,368$ \\
            NQ & $15,000$ & $8,006$\\
            NarrativeQA & $7,000$ & $--$ \\
            SQuAD & $67,00$ & $--$ \\
            TruthfulQA & $817$ & $--$ \\
            \hline
        \end{tabular}
    \caption{Number of examples in each Dataset used in the fine-tuning experiments.}
    \label{tab:ft data collection}
    \vspace{-5mm}
\end{wraptable}
subset from the training set to avoid overlapping. The exact number of entries in each train and test set is detailed in Table \ref{tab:ft data collection}.

We use the dataset-provided documents as $d_{gold}$ for each data entry. To obtain $d_{random}$ we sample the context of different entries within the same dataset, to make sure the distributions of $d_{random}$ and $d_{gold}$ are roughly similar.

\noindent\textbf{Metrics}
\quad We use the ground-truth coverage as our evaluation metric, considering that the answers of QA tasks are relatively short, while the generation length of the model is sometimes hard to limit. 

\noindent\textbf{Implementation Details}
\quad We select Llama-2-7b~\cite{Touvron2023Llama2O} as the base model. For efficiency, we use LoRA~\cite{Hu2021LoRALA} and int8 quantization during training.
The prompt templates used for fine-tuning and evaluation mainly follow ~\citet{Lin2023RADITRD}. We train our generator for 3 epochs and constrain the maximum length of the sequence to 1600, using a batch size of 4 and a learning rate of 5e-5.  
During testing, we use a zero-shot setting. 

\noindent\textbf{Detailed Results}
\quad Table \ref{tab:ft result} shows our evaluation results on each dataset.

\begin{table*}[t]
    \renewcommand{\arraystretch}{1.5}
    \centering
    \setlength{\tabcolsep}{10pt} 
    \small
    \begin{tabular}{l   p{11cm}}
\hline
\textbf{[Instruction]} & Please generate ten descriptions for the continuation task. \\
\hdashline
\textbf{[Context]} & 
For example:

1.``French.Washington played a crucial role in the American Revolutionary War, leading the Continental Army against the British.'' Please continue writing the above paragraph.

2.``The discovery of the double helix structure of DNA by James Watson and Francis Crick revolutionized the field of genetics, laying the foundation for modern molecular biology and biotechnology.'' Please continue by discussing recent developments in genetic research, such as CRISPR gene editing, and their potential ethical implications. \\
\hline
    \end{tabular}
    \caption{\label{tab:task generate}Template for generating task classification data.}
\end{table*}

\subsection{Experimental Details of Comprehensive Evaluation}
\label{Evaluation Appendix}

\noindent\textbf{Tasks and Datasets}
\quad We conducted extensive experiments across various NLP tasks and datasets to assess the performance of RAG systems. Specifically:
(1) \textbf{Commonsense Reasoning}: We evaluated on MMLU~\cite{Hendrycks_Burns_Basart_Zou_Mazeika_Song_Steinhardt_2020}, ARC-Challenge~\cite{Clark2018ThinkYH}, and OpenbookQA~\cite{Mihaylov_Clark_Khot_Sabharwal_2018} datasets.
(2) \textbf{Fact Checking}: Our evaluation encompassed the FEVER~\cite{Thorne2018FEVERAL} and PubHealth~\cite{Zhang2023InterpretableUL} datasets.
(3) \textbf{Open-Domain QA}: We assessed on NQ~\cite{Kwiatkowski2019NaturalQA}, TriviaQA~\cite{Joshi2017TriviaQAAL}, and WebQuestions~\cite{Berant_Chou_Frostig_Liang_2013} datasets.
(4) \textbf{MultiHop QA}: Our evaluation included the HotPotQA~\cite{yang2018hotpotqa}, 2WikiMultiHopQA~\cite{Ho2020ConstructingAM}, and MuSiQue~\cite{Trivedi_Balasubramanian_Khot_Sabharwal_2022} datasets. For MuSiQue, we followed the approach outlined in~\cite{Press_Zhang_Min_Schmidt_Smith_Lewis_2022} and focused solely on answerable 2-hop questions.
(5) \textbf{Medical QA}: We also assessed on the PubMedQA~\cite{Jin2019PubMedQAAD} dataset.
In each dataset, we randomly sub-sample 500 entries from the test set for our experiments. For datasets without test set, we use develop set instead.  

To assess RAG capabilities, we evenly collect a total of 500 entries from NQ, TriviaQA, HotPotQA, 2WikiMultiHopQA and MuSiQue. 
Each entry is a ``question, gold document, gold answer'' triple.

\noindent\textbf{Metrics}
\quad We use token-level F1 score and EM score for Open-Domain QA and MultiHop QA tasks, and accuracy for others. 
We use a more lenient EM score, which evaluates performance based on whether the model generations include gold answers instead of strictly exact matching~\cite{asai2023self}.

Towards RAG capabilities evaluation, we adopt four metrics from RAGAs, including \textbf{Faithfulness}, \textbf{Context Relevancy}, \textbf{Answer Relevancy}, and \textbf{Answer Correctness}. 
Faithfulness measures how factually consistent the generated answer is with the retrieved context. An answer is considered faithful if all claims made can be directly inferred from the provided context.
Context Relevancy evaluates how relevant the retrieved context is to the original query.
Answer Relevancy assesses the pertinence of the generated answer to the original query.
Answer Correctness involves the accuracy of the generated answer when compared to the ground truth.
For example, Context Relevancy is calculated from the proportion of sentences within the retrieved context that are relevant for answering the given question to all sentences:
 
\begin{equation}
    context \; relevancy = \frac{\left | S \right | }{\left | Total\right | } 
\end{equation}
where $\left | S \right |$ denotes the number of relevant sentences, $\left | Total \right |$ denotes the total number of sentences retrieved.
All these metrics are evaluated using the RAGAs framework, with GPT-4 serving as the judge. 

Additionally, we compute the cosine similarity between the retrieved document and the gold document as \textbf{Retrieval Similarity}.
The retrieved document and gold document are fed into an embedding model, then the resulting embeddings are used to compute the cosine similarity.

\noindent\textbf{Implementation Details}
\quad For Open-Domain QA and MultiHop QA datasets, we set the generation model's maximum new token number to 100 tokens.
For other datasets, we set it to 50 tokens. To deal with excessively long retrieved documents, we truncated the documents to 2048 words when evaluating RankLLaMA and LongLLMLingua.

For all datasets, we use greedy decoding during generation. To better compare the capabilities of different RAG modules, we adopt the 0-shot evaluation setting,
i.e., no in-context examples are offered.
In the multiple choice and fact checking tasks, answers generated by the model may take a variety of forms (e.g., ``the answer is A'' instead of ``A''). 
Therefore, we preprocess the responses generated by the model, applying regular expression templates to match them with gold labels.



\end{document}